\documentclass{article}

\usepackage{geometry}
\usepackage{microtype}
\usepackage{graphicx}
\usepackage{subfigure}
\usepackage{booktabs} 

\usepackage{hyperref}

\usepackage{placeins}

\usepackage[toc,page]{appendix}
\usepackage[font=small,labelfont=bf]{caption}
\usepackage{float}

\title{}

\title{Convolutional Neural Networks In Convolution}
\author{Xiaobo Huang \\ RDF International School \\ hxb@mws.site}

\begin{document}

\maketitle

\begin{abstract}
    Currently, increasingly deeper neural networks have been applied to improve their accuracy.
    In contrast, We propose a novel wider Convolutional Neural Networks (CNN) architecture, motivated by the Multi-column Deep Neural Networks\cite{ciregan2012multi-column} and the Network In Network(NIN)\cite{lin2014network}, aiming for higher accuracy without input data transmutation.

	In our architecture, namely ``CNN In Convolution"(CNNIC), a small CNN, instead of the original \emph{generalized liner model}(GLM) based filters, is convoluted as kernel on the original image, serving as feature extracting layer of this networks.
	And further classifications are then carried out by a global average pooling layer and a softmax layer.

    Dropout and orthonormal initialization are applied to overcome training difficulties including slow convergence and over-fitting. Persuasive classification performance is demonstrated on MNIST\cite{lecun-mnisthandwrittendigit-2010}.
\end{abstract}

\section{Introduction}

        CNN\cite{lecun1998gradient-based-mnist} is one of the classical architectures that reaches a decent performance on object recognition tasks, and deep CNNs\cite{he2016deep} have been taken as conventional architectures approaching state of art performance in object recognition tasks.
        
        The depth of CNN, the numbers of convolutional layers in a network, are usually directly and positively related to its performance.
        Thus, increasing work \cite{Krizhevsky2012ImageNetCW} \cite{simonyan2015very} \cite{szegedy2015going} \cite{szegedy2016rethinking} has been performed on methods of approaching deeper network. 
        While much research has been conducted to boost the depth of the network, meanwhile, the resistance encountered when creating a deeper network, like exploding or vanishing gradient problem, intensives. The conventional solution using Deep Residual Structure\cite{Kaiming2015resnet} addressing the fore-mentioned problems, implicitly breaks a deep network into the addition of multiple shallower substitutes. We thus predict that a wider approach with CNN may as well lead to improved discriminability without burdens of deeper structures. 

        Ensemble-based classifiers\cite{rokach2010ensemble-based}, the foundation of wider networks, combines an ensemble of weighed individual sub-classiﬁers trained with differently manipulated data-sets to acquire a performance over any individual classifier inside the ensemble.

        Existing research\cite{ciregan2012multi-column} has been performed on ways making CNN wider via using ensemble-based CNN with varied inputs pre-processed using data augmentation inspired by micro-columns of neural in the cerebral cortex\cite{Jones2000Microcolumns}.

        In this work, we adopt both the advantages of ensemble-based classifiers and the input pre-processing method of strided convolution inspired by NIN\cite{lin2014network}.
        The novel elements in our model include feeding different part of the data to the classifiers using strided convolution while all classifiers share the same set of weight and the output of each sub-classifier polls to generate the final classification results.
        The weight-sharing and non-recurrent structure allow the architecture to own less weights\footnote{116,3980 in total for CNNIC-2} and better parallelizing ability.
        Our novel way of ensembling contains much less parameters and smaller classifiers, having higher speed as well as state of art performance.
        What's more, our work also proved that for a fix number of parameters, wider architectures, comparing to those that are deeper, are also suitable for improving the performance in object recognition tasks, and based on our experiments, combining wider architecture with deeper ones may be futurous for further research.

\section{Previous Work}
        \subsection{Ensemble Classifiers}
        There are two main categories of ensemble learning methods: dependent methods and independent methods, which only the latter is relevant to our discussion.
        It has been proven\cite{Krogh1994NeuralNE} that, the improved performance comes from the variety of the ensemble, which to the foundation be the ambiguity between the output of different sub-classifiers. 
        Assuming all sub-classifiers having the same structure, the only source of variety would be the differently manipulated input data. The specific input data pre-processing methods vary among different architectures.
    \subsection{Convolution Neural Network(CNN)}
        CNNs\cite{lecun1998gradient-based-mnist} are chiefly constructed with convolutional layers allowing the network to extract features in an translation-invariant manner using learnable filters with much less weight. 
        
        The traditional filters in the convolutional layers are Generalized Linear Models(GLMs) which calculates the output using plain convolution operation. 
        The outputs of the classic convolutional layer using ReLU can be represented as follows:
        $$f_{i,j,k}=max(w^T_k x_{i,j} + b_k, 0)$$
        , where $(i, j)$ is the pixel index in the feature map, $x_{i,j}$ stands for the input patch centered at location$(i, j)$, and $k$ is used to index the channels of the feature map.
        
        The classical convolutional layer only uses simple learnable linear filters to pick up raw shapes on the feature map, thus to identify more complex features in an image usually requires massive stacking of layers. 

    \subsection{Network In Network(NIN)}
        NIN\cite{lin2014network} is a modified version of normal CNN architecture using small Multi-layer Perceptron (MLP) to replace the GLMs as a kernel to make the kernel structure more complex, enhancing their the ability to identify much more complex shapes.
        
        The outputs of the convolutional layer in NIN are as follows:
        $$f_{i,j,k}=mlp(x_{i,j}, W_k)$$
        Here $mlp(x,W)$ is the outputs of a micro neural network with the inputs $x$ and weights $W$.
        
        This architecture has significantly improved performance because the boosted fitting ability and generalization ability by replacing the convolutional filters with a MLP operation.
        
    \subsection{Global Average Pooling(GAP)}
        NIN also adopted a new output layer called Global Average Pooling to replace the traditional fully connected layers used in CNN because the latter are prone to over-fitting.
        GAP takes the average of each feature map, and the resulting vector is fed directly into the final softmax layer.
    \subsection{Multi-column Deep Neural Network}
        Multi-column Deep Neural Network\cite{ciregan2012multi-column} is another architecture assembling multiple DNNs seeking enhanced performance. Being inspired by the columns structure of cerebral cortex, Multi-column Deep Neural Network groups several weighted DNN, called column, and then averages the classification results of multiple columns. 
        
        As an advantage of being an independent ensemble framework, its multi-columed structure allows it to be trained or used in parallelizing manner, boosting its speed. 
        
        What is worth noticing in this model is that the input images for different columns are preprocessed by different inducers to increase the ambiguity of sub-DNNs. The final predictions are then obtained by averaging individual predictions of each DNN. 
        
\section{Structure}
    \subsection{Strided Convolution}
        Strided Convolution a widely adopted mean to simplify convolution operations. 
        In traditional convolution, the filters slide in fixed steps of one pixel; 
        for strided convolution, the filters slide across multiple pixels, reducing the size of the output feature map.
        Strided convolution is has been widely adopted to replace the combination of convolutional layer and pooling layer, leading to faster computation.
    \subsection{Small CNN}
        Abducting from both ``Network In Network"\cite{lin2014network} and ``Maxout Networks"\cite{Goodfellow2013Maxout}, the classification ability of a CNN improves by increasing the complexity of the filters, thus we choose classical CNNs as our filters, in respect to their complexities out-weighting the other choices. 

        The small CNNs used are deliberately designed to be shallower, containing only three or two convolutional layers and two fully-connected layers, for the sake of saving computational resources meanwhile preserving enough complexity. 
        
        We carried out two small CNN architectures, where the former has two convolutional layers and the latter has three, as shown in Table 1.
        By using normal CNN architectures, we hope to prove the performance increase comes from the overall architecture rather than any specific feature used in small CNNs themselves.
        
        \begin{table}[H]
    	    \begin{center}
    		    \begin{tabular}{ccc}
    		\hline  Layer type & CNNIC-3 & CNNIC-2 \\ \hline
        			   Conv    & 5x5@32  & 5x5@64  \\
        			   Conv    & 5x5@64  &  \\
        			Avg\_pool  & 2x2(D)  & 2x2(D)  \\
        			   Conv    & 5x5@64  & 5x5@64  \\
        			Avg\_pool  & 2x2(D)  & 2x2(D)  \\
        			    FC     & 1024(D) & 1024(D) \\
        			 Softmax   &  10(D)  &  10(D) \\ \hline
        		\end{tabular}\\
        		("D" label indicates dropout applied on layer output)
        	\end{center}
        	\caption{Alternative Architectures Adopted and Tested as Small CNN}
        \end{table}

    \subsection{CNN In Convolution(CNNIC)}
        \begin{figure*}
        	\begin{center}
        		\includegraphics[width=\linewidth]{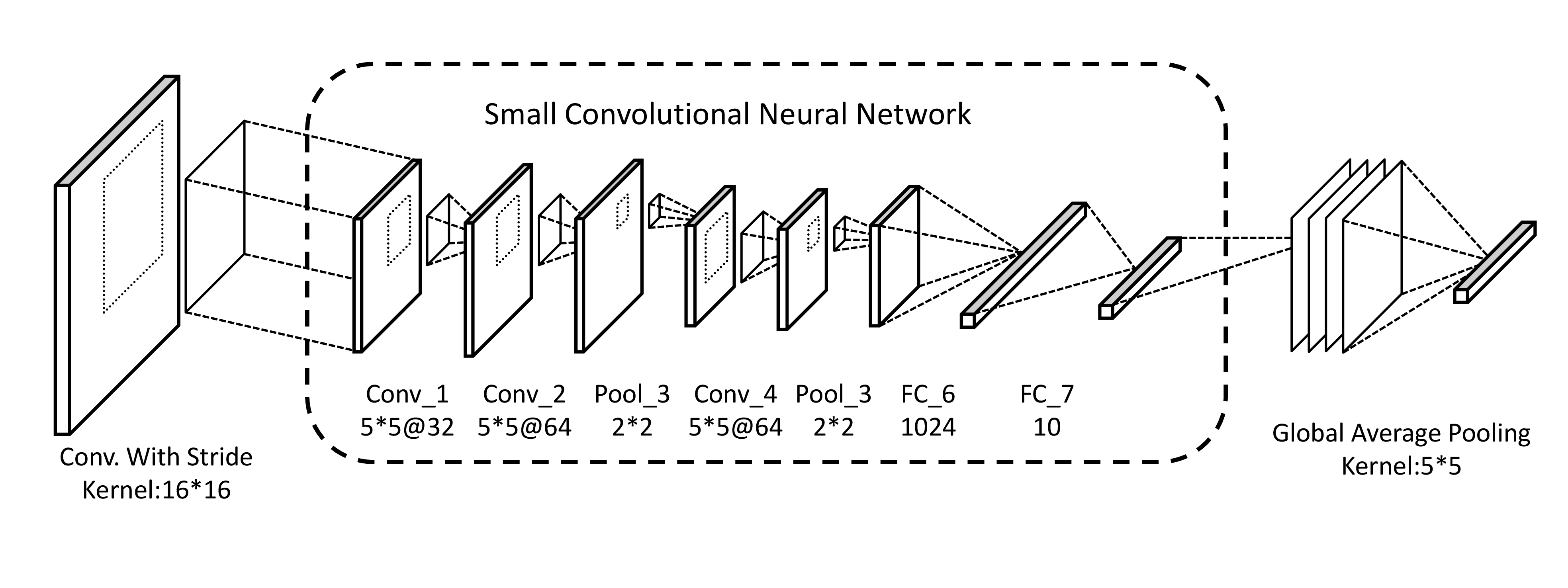}
        	\end{center}
    	    \caption{The structure of the CNNIC framework.}
        	\label{fig:short}
        \end{figure*}
        
        With the structure of the kernel defined, the overall structure of CNN In Convolution could be summarized into a convolutional layers whose filters are replaced with a fixed number of small CNNs (one in our experiments), followed with a global pooling layer which has been proven by practice to have better generalization ability and could prevent over-fitting.
        The results from the small CNN kernels on different areas of convolution are then averaged and soft-maxed for the final classification answer, as illustrated in Figure 1 for a more intuitive view. 
        
        In the light of ensemble learning, this framework could be considered as an independent ensemble of weight-sharing classical CNNs as base-classifiers. The input image is cropped into evenly strided pieces and the input data of each classifier is chosen from one of the pieces. 

\section{Experiments}
    \subsection{Overview}
        We bench-marked the performance of both CNNIC-2 and CNNIC-3 within the MNIST\cite{lecun-mnisthandwrittendigit-2010} data-set. 
        Dropout\cite{srivastava2014dropout:}, a widely used approach to prevent over-fitting is adopted in our experiment to reduce generalization error, where the dropout probabilities are set uniformly to $40\%$. 
        
        During training, experiment shows that Adam optimizer\cite{Kingma2014Adam} works the best on boosting both convergence speed and accuracy. We also found the model sensitive to initial learning rate for that a lower rate(e.g. $10^{-5}$ when using Adam-optimizer) causes the architecture to under-fit the training data while a big learning rate(e.g. $0.003$ when using Adam-optimizer) causes it to over-fit the training data.
        
        During training, the model is observed to suffer from convergence difficulties. Still, adopting an initial learning rate of $10^{-3}$ with attenuation enables accessibly training of the model.
    
        Our Experiment also indicates that architecture with more small CNNs generally performs significantly worse by over-fitting simple datasets. 
        
        All of our experiments are performed on one NVIDIA GTX 1060 6GB, based on Tensorflow. The corresponding code could be found here\footnote{\url{https://github.com/MyWorkShop/Convolutional-Neural-Networks-in-Convolution}}.

    \subsection{MNIST}
        \begin{figure}[!ht]
        	\begin{center}
        		\includegraphics[width=0.6\linewidth]{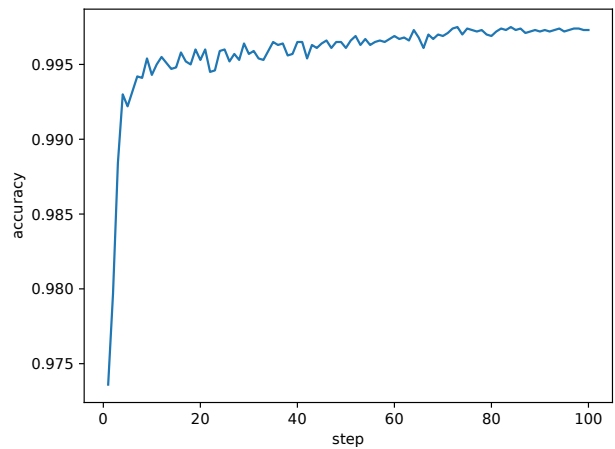}
        	\end{center}
    	    \caption{Accuracy change by steps of experiment on MNIST}
        	\label{fig:training}
        \end{figure}
        MNIST\cite{lecun1998gradient-based-mnist} is a small handwritten digits data-set, which has a training set of 60,000 examples, and a test set of 10,000 examples.
        We use it as the main data-set in architecture adjustment. The training status and final accuracy are displayed in figure \ref{fig:training} and table \ref{table:acc} respectively. 
        \begin{table}[H]
	        \begin{center}
	        	\begin{tabular}{cc}
	        		                 Method                   & Test Error \\ \hline
	        		Maxout Network\cite{Goodfellow2013Maxout} &   0.47\%   \\
	        		        NIN\cite{lin2014network}          &   0.45\%   \\
	        		            \textbf{CNNIC-2}              &   0.38\%   \\
	        		          MIM\cite{liao2016on}            &   0.35\%   \\
	        		          \textbf{CNNIC-3}              &   0.33\%   \\
	        		    RCNN-96\cite{liang2015recurrent}      &   0.31\%   \\
	        		   MCDNN\cite{ciregan2012multi-column}    &   0.23\%
	        	\end{tabular}
	        \end{center}
        	\caption{Test set error rates for MNIST of different architectures.}
        	\label{table:acc}
        \end{table}


    \subsection{More than one Small-CNNs}
    
    Architectures with more than one Small-CNNs inside the CNNIC layer are also tested in or experiment, however, limited by computational resources, only on low resolution datasets. 
    
    We use this simple formula to measure the over-fitting situation of the network:
        $$\mathit{O}=(\frac{E_{\mathit{train}}}{N_{\mathit{train}}}-\frac{E_{\mathit{test}}}{N_{\mathit{test}}})$$, 
    where $O$ is the over-fitting index, $E$ is the total error and $N$ is the size of the corresponding batch.
    
    We observe that the over-fitting index increases drastically as more than one Small-CNNs are employed in training for simple data-sets like MNIST. We believe using more than one Small-CNNs is potentially exploitable on more complex datasets, but are unable to test it limited by computational resources.

\section{Discussion}

    \subsection{Intrepting the Effectiveness of CNNIC}
    For an ensemble of classifiers, it has been proven\cite{Krogh1994NeuralNE} that:
        $$E = \overline{E} - \overline{A}$$
    , where $E$ is the generalization error of the ensemble, 
    $\overline{E}$ is the average generalization errors of the individual networks, and $\overline{A}$ is the average of their ambiguities, in another word, variances among members of the ensemble. The variances among each classifier are key to low generalization error of the ensemble. 
    A CNNIC network, in light of ensemble learning, is an ensemble of weight-sharing classical CNNs. Different from traditional ensembling methods whose ambiguity came from weight differences of base classifiers, the ambiguity of CNNIC comes from different inputs, reducing total amount of parameters without the expense of ambiguities. 
    
    Besides encouraging CNNIC to preserve ambiguity among each base-classifiers, this setup also encourages more efficient weights usage. By squeezing multiple base-classifiers into a small shared set of weight, reuse of low-level feature extraction kernels is almost a certainty. At the same time, the number of weights for each base-classifier, rather than hard-limited in traditional ensemble models, could be allocated as needed with gradient descent in CNNIC. 
    
    Interpreting CNNIC via data augmentation, however, may not be appropriate. While the input image is cropped into smaller sections for training, the performances of each individual classifier are poor. Also, passing location information via CoordConv\cite{Liu} slightly increase network performance. These observations indicate the superior performance of CNNIC comes from its ensemble structure, rather than training inner CNNs with augmented datasets. 
    
    \subsection{Dropout and Co-adaption}
    
    According to Hinton\cite{Hinton2012Improving}, Dropout is a mean serves to prevent over-fitting and possible development co-adaption, in another word, dependency, between different portion of feature detectors by separating them. From our experiment we observe that adding dropout will significantly enhance performance of our network, even after having ruled out its effect preventing over-fitting, a well-defined issue.
    
     It has been purposed\cite{Nannen03paradox} that the complexity of a model $M$ could be measured by its Minimum Description Length(MDL), in another word, the total Shannon Entropy of it $n$ parameters $\theta_n$,
        $$E(M)=\sum Pr(\theta_n) \log_2(Pr(\theta_n))$$
        , where in randomly initialized models could be simplified as 
        $$E(M)=n\log(Pr(\frac{1}{n}))$$
    
    It's also purposed\cite{Hastie01a} that the trade-off between training set accuracy and over-fitting could be view as a method of information compression from the input $\textbf{X}$ and parameters $\theta$ to the desired output $\textbf{y}$. And the total Description Length of the model $M$ is 
        $$D(M) = -\log(\textbf{y}|\theta,M,\mathbf{X}) - \log(\theta|M)$$
        
    Our network should find a balance by finding $\arg \max_M D(M)$, and over-fitting occurs when the price of accuracy, $\log(\theta|M)$, gets too high. 
    
    Thus, over-fitting problem should always be addressable by simplifying the model, in turn, decreasing $\log(\theta|M)$. However, while higher dropout rate continues to enhance performance, simplifying the model at the same time does the opposite. From the theorem above that reducing the MDL of the model is not helping the performance, thus there exists not significant over-fitting in the current model. This could also be proved by the fact that CNNIC-3 outperforms the CNNIC-2 model.

    What left in dropout that may have positive impact on model performance is the prevention of co-adaption between different portion of the network. we could not find a mathematical definition of co-adaption, nor the existence of a full study. CNNIC may be used for future study of the effect of co-adaption where the prevention of over-fitting by dropout is eliminated. 

\section{Conclusion}
We purposed a novel artificial neural network architecture, CNNIC, for image classification tasks, by replacing the convolution operations in traditional CNN with a smaller CNN, followed with global average pooling. We demonstrated state-of-art performance on MNIST dataset, achieving so with smaller parameter count comparing to other architectures like multi-column DNN. We explained the behaviors of the network using principle of ensemble learning, suggesting a novel way creating ambiguity in ensemble classifiers without inflation of parameter counts by manipulating inputs of a set of weight sharing classifiers. 

\section*{Acknowledgements}

    Tao Huang, also the author of this paper, refused to acknowledge his name in the author section. We sincerely thank him for his significant contribution in both the writing and founding of the paper.


\bibliography{example_paper}
\bibliographystyle{plain}
\nocite{*}

\end{document}